\documentclass[letterpaper,10pt]{article}
\usepackage{amsmath,epsfig}
\usepackage[preprint]{spconfa4}
\let\OLDthebibliography\thebibliography
\renewcommand\thebibliography[1]{
  \OLDthebibliography{#1}
  \setlength{\parskip}{0pt}
  \setlength{\itemsep}{0pt plus 0.3ex}
}

\usepackage{amsmath,amssymb,amsfonts}
\usepackage{algorithmic}
\usepackage{graphicx}
\usepackage{textcomp}
\usepackage{xcolor}
\usepackage[noadjust]{cite}
\usepackage{float}
\usepackage{makecell}
\usepackage{multirow}

\begin{document}
\title{Warwick Image Forensics Dataset for Device Fingerprinting In Multimedia Forensics}
\name{Yijun Quan$^{\ast}$, Chang-Tsun Li$^{\dagger}$, Yujue Zhou$^{\ast}$ and Li Li$^{\ddagger}$}
\address{$^{\ast}$Department of Computer Science, University of Warwick, UK\\
 E-mail: \{y.quan, y.zhou.34\}@warwick.ac.uk;\\ $^{\dagger}$School of Information Technology, Deakin University, Australia\\ E-mail: changtsun.li@deakin.edu.au;\\$^{\ddagger}$ School of Computer Science and Technology, Hangzhou Dianzi University, China  \\ E-mail:  lili2008@hdu.edu.cn}

%

\maketitle
\begin{abstract}
Device fingerprints like sensor pattern noise (SPN) are widely used for provenance analysis and image authentication. Over the past few years, the rapid advancement in digital photography has greatly reshaped the pipeline of image capturing process on consumer-level mobile devices. The flexibility of camera parameter settings and the emergence of multi-frame photography algorithms, especially high dynamic range (HDR) imaging, bring new challenges to device fingerprinting. The subsequent study on these topics requires a new purposefully built image dataset. In this paper, we present the Warwick Image Forensics Dataset, an image dataset of more than 58,600 images captured using 14 digital cameras with various exposure settings. Special attention to the exposure settings allows the images to be adopted by different multi-frame computational photography algorithms and for subsequent device fingerprinting. The dataset is released as an open-source, free for use for the digital forensic community.
\end{abstract}

\section{Introduction}

Image device fingerprinting is an important topic in multimedia forensics. It allows forensic investigators to establish an image's history, identify the source device and authenticate the content. Sensor Pattern Noise (SPN) \cite{1634362}, as its name suggests, is a noise intrinsically embedded in images, primarily due to Photo Response Non-Uniformity (PRNU). Such an intrinsic property makes SPN a popular candidate for device fingerprinting and many researches are done on SPN-based source camera identification\cite{1634362,li2010source}, tampering localization \cite{chen2008, 5934587} and source camera clustering \cite{caldelli2010fast, Li2017, 8753503}. Public datasets like Dresden Image Dataset \cite{gloe2010dresden} and VISION Image Dataset \cite{Shullani2017}, which can be used as benchmarking platforms, are very important for the study of device fingerprint analysis and the development of relevant techniques.\\
\indent As the digital forensic community is gaining more understanding of image device fingerprinting, digital and computational photography has undergone huge development as well. Driven by the need for consumer-level devices to produce better images, we witness significant advances in both hardware and software development. As far as hardware is concerned, the improvement in the design of electronic components like complementary metal-oxide-semiconductor (CMOS) brings better noise immunity. Such improvements allow cameras to have greater flexibility in camera parameter settings, especially for using high signal gain (commonly known by the name of \textit{ISO speed} in photography) without introducing too much noise to images. Thus, digital photography becomes more versatile under different lighting conditions and can be used for high-speed photography. In addition, the ever-increasing computational power of consumer-level mobile devices brought by the improvement in hardware allows more sophisticated computational photography algorithms to be processed in real-time. Among these algorithms, merging multiple time-sequential image frames is a very popular computational photography strategy used by consumer-level devices, especially for \textit{high dynamic range} (HDR) imaging \cite{reinhard2010high}. By processing a burst shots of images, the resultant image can be of higher dynamic range, less noisy and often aesthetically more appealing. Thus, the HDR imaging mode has received great popularity and become available in most mobile imaging devices.

While the above mentioned improvements are greatly appreciated by the users, new challenges are faced by existing SPN-based device fingerprinting methods. Often, existing SPN-based device fingerprinting methods are working on the correlation between the noise residuals extracted from the images. The intra-class correlations (the correlations between noise residuals of images from the same source device) can be greatly affected by images' ISO speeds and the alignment operation used in multi-frame computational photography algorithms. This results in compromised forensic accuracy when running existing SPN-based methods on these images. Thus, insightful investigations are required to understand the problems behind and develop effective forensic methods accordingly. However, the images of the existing datasets in the public domain are not purposefully collected to help answer these problems. Therefore, we have built a new dataset called \textit{Warwick Image Forensics Dataset}, which can not only serve the same purposes as the existing datasets, but also includes images with their source cameras working in different exposure settings. It is intended to pave the way for finding methods to deal with the impact on the accuracy of device fingerprinting due to exposure parameter settings and multi-frame computational photography algorithms.\\
\indent The rest of the paper is organized as follows. In the next section, related work, including existing forensic datasets, will be discussed. The details of the Warwick Image Forensics Dataset are presented in Section \ref{sec:dataset} and experimental evaluations are carried out in Section \ref{sec:exp}. A conclusion is given in Section \ref{sec:con}.
\section{Related Work}
\subsection{ISO Speed's Impact On SPN-Based Digital Forensics}
SPN, as a fixed pattern noise, primarily arises from PRNU. \cite{chen2008} considers an image $\textbf{I}$ with a sensor output model as:
\begin{equation}
  \mathbf{I} = g^\gamma \cdot[(\mathbf{1}+\mathbf{K})\mathbf{Y} + \mathbf{\Lambda}]^\gamma + \mathbf{\Theta}_q 
\end{equation}
where $g$ is the camera gain, $\gamma$ is the gamma correction factor and $\mathbf{Y}$ is the scene light intensity. The model considers two major noise terms, represented by $\mathbf{\Lambda}$ and $\mathbf{\Theta}_q$, respectively. $\mathbf{\Lambda}$ is a combination of noise sources including dark current, shot noise and the read-out noise. $\mathbf{\Theta}_q$ represents the quantization noise. The PRNU term of our interest is represented by $\mathbf{K}$, showing the non-uniform response to the scene light intensity $\mathbf{Y}$. The model is simplified in \cite{chen2008} by exploiting the Taylor expansion of the gamma correction and can be written as:
\begin{equation}
 \mathbf{I} \doteq \mathbf{I}^{(0)} + \mathbf{I}^{(0)} \mathbf{K} + \mathbf{\Theta}
\end{equation}
with $\mathbf{I}^{(0)} = (g\mathbf{Y})^\gamma$, being the sensor output in the absence of noise, and $\mathbf{\Theta} = \gamma \mathbf{I}^{(0)}\mathbf{\Lambda}/\mathbf{Y} + \mathbf{\Theta}_q$, being a complex of PRNU-irrelevant random noise components. Written in this form, the PRNU component $\mathbf{I}^{(0)} \mathbf{K}$ is a multiplicative term with the noise free image $\mathbf{I}^{(0)}$. However, the role of camera gain, $g$, in the sensor output model can be easily ignored. Given similar $\mathbf{I}^{(0)}$ from different images, the size of $\mathbf{\Theta}$ would differ with different camera gain $g$ as higher $g$ requires less input intensity $\mathbf{Y}$ to produce the same output signal $\mathbf{I}^{(0)}$. As $\mathbf{\Theta} = \gamma \mathbf{I}^{(0)}\mathbf{\Lambda}/\mathbf{Y} + \mathbf{\Theta}_q$, a smaller $\mathbf{Y}$ will induce more PRNU-irrelevant noise in an image's noise residual. Because SPN is often estimated as the noise residual of an image, the addition of SPN-irrelevant images will make this image's noise residual less correlated with noise residuals extracted from other intra-class images.

With the above relationship in mind, in \cite{8451688}, the authors empirically show that given similar contents in images taken with different ISO speed settings, the intra-class correlation distributions can vary according to ISO speeds, which directly control the camera gain $g$. This results in higher error rates in source camera identification for images of higher ISO speeds. Due to this phenomenon, \cite{8451688} suggests that camera exposure parameters like ISO speed should be considered from a forensic perspective. It is also suggested that the construction of forensic image datasets should include images of different exposure parameter settings, which can also be beneficial for studies in steganalysis.
\subsection{High Dynamic Range Imaging}
HDR images can capture more details from scenes compared to standard dynamic range (SDR) images and hence receive much attention from computational photography researchers. From the early works in \cite{mann1994beingundigital, debevec2008recovering} to the more recent works like HDR+ \cite{hasinoff2016burst} and deep neural network based methods \cite{eilertsen2017hdr}, different HDR imaging techniques are developed to allow them to be used under different conditions. Despite the differences, these methods also share a few things in common, which make HDR images a hard subject in general for SPN-based device fingerprinting. For most HDR imaging algorithms, conventional exposure methods of taking a set of time-sequential images are often used, despite some methods have images with the same exposure time and some others use images with different exposure time. A radiance map can be reconstructed from a set of time-sequential images and provides a larger dynamic range than single exposure images. However, as it is almost impossible to avoid object or camera motion during the capturing process of the time-sequential image sets, the reconstruction of the radiance map usually involves pixel-wise alignment to compensate the object motions across different image frames to avoid motion blurring. Such an operation will mix the SPN signal from different pixel and cause misalignment between the SPN embedded in the resultant HDR images and reference SPN extracted from single exposure images taken by the same camera. Due to such misalignment, intra-class SPN pairs will be less correlated and cause difficulty in SPN-based provenance analysis.

In addition to the misalignment problem, tone mapping is another operation commonly used in HDR algorithms, which can cause trouble for existing SPN-based forensic methods. Tone mapping is used to reconstruct a color image from a radiance map. Each implementation of different HDR algorithms may have its unique tone mapping curve and on top of that, different tone mapping curves can be applied either globally or locally on the same image. As SPN-based forgery localization methods often use a content dependent correlation predictor to estimate the block-wise intra-class correlations to discover pixels with its SPN absent, without the prior knowledge of the tone mapping curve, reliable predictions from the correlation predictor can hardly be expected. These problems require specific adjustment for existing SPN-based methods to make them effective on HDR images.
\subsection{Existing Public Image Datasets}
As a rapidly developing topic, device fingerprinting draws many researchers' attention and several image datasets are constructed over the years to facilitate the researches. One of the earliest image datasets adopted for device fingerprinting is the Uncompressed Colour Image Dataset (UCID)\cite{schaefer2003ucid}. From then on, more dedicated image datasets for provenance analysis are constructed. Notably, the Dresden Image Dataset \cite{gloe2010dresden}, RAISE dataset \cite{dang2015raise} and VISION dataset \cite{Shullani2017} are three datasets widely used for benchmarking in device fingerprinting. Each dataset consists of a large number of high resolution images from multiple devices, either digital cameras or smartphone cameras. More recent datasets like the SOCRatES \cite{galdi2019socrates} and DAXING datasets \cite{8760241} feature images from a vast number of source devices (103 smartphone cameras from SOCRatES and 90 smartphone cameras from DAXING dataset). Despite the images from these datasets show good diversity and heterogeneity in terms of contents, all the above mentioned datasets focus on SDR images only and the diversity in camera exposure parameter settings was not given adequate consideration during the construction of these datasets.

The `HDR dataset' from \cite{shaya2018new} is the first forensic dataset featuring HDR images. The images in this dataset are taken with 23 smartphone cameras and for each scene included in this dataset, both a SDR image and a HDR image are provided. The images are taken under three different conditions: taken from the tripod, by the hand and by a shaky hand. Despite \cite{shaya2018new} featuring both SDR and HDR images, its real contribution of the image pairs towards the understanding of HDR images' impact on source device identification is limited. Firstly, the SDR images included in the dataset are not the SDR images used for the construction of the HDR images. As a result, these pairs may not best reflect the impact of HDR algorithms on device fingerprints in SDR images. Secondly, as the HDR images in this dataset are generated directly from the smartphones, the coverage of different implementations of HDR algorithms are confined by the choice of smartphones included in this dataset. As the development of new HDR algorithms continues, research findings stemmed from this dataset are unlikely to be applicable to other HDR images produced by future algorithms. Acknowledging this problem, our Warwick Image Forensics Dataset takes the flexibility of generating HDR images using different implementations of HDR algorithms into account as we shall see from the following section.
\section{Dataset Details}
In this section, we present the details of our Warwick Image Forensics Dataset.
\label{sec:dataset}
\begin{table*}
\begin{center}
\caption{Details of the cameras presented in Warwick Image Forensics Dataset}
\label{tab:cameras}
\begin{tabular}{c c c c c c c c c}
\hline 
No. & Camera & Resolution & \makecell{Sensor\\ Format} & Sensor Dimensions  & CFA Type & Lens\\ \hline
1 & Canon EOS 6D & $3648 \times 5472$ & 35 mm & 35.8 $\times$ 23.9 mm$^2$ & Bayer Filter & Interchangeable\\
2 & Canon EOS 6D Mark II & $4160 \times 6240$ & 35 mm & 35.9 $\times$ 24 mm$^2$ & Bayer Filter & Interchangeable\\
3 & Canon EOS 80D & $4000 \times 6000$ & APS-C & 22.5$\times$ 15 mm$^2$ &  Bayer Filter & Interchangeable\\
4 & Canon EOS M6 & $4000 \times 6000$ & APS-C & 22.3$\times$14.9 mm$^2$ & Bayer Filter & Interchangeable\\
5 & Fujifilm X-A10\_1 & $3264 \times 4896$ & APS-C & 23.6$\times$15.6 mm$^2$ & Bayer Filter & Interchangeable\\
6 & Fujifilm X-A10\_2 & $3264 \times 4896$ & APS-C & 23.6$\times$ 15.6 mm$^2$ & Bayer Filter & Interchangeable\\
7 & Nikon D7200 & $4000 \times 6000$ & APS-C & 23.5$\times$ 15.6 mm$^2$ & Bayer Filter & Interchangeable\\
8 & Panasonic Lumix DC-TZ90\_1 & $3888 \times 5184$ & 1/2.3''& 6.16 $\times$ 4.62 mm$^2$& Bayer Filter & Fixed\\
9 & Panasonic Lumix DC-TZ90\_2 & $3888 \times 5184$ & 1/2.3''&6.16$\times$ 4.62 mm$^2$&  Bayer Filter & Fixed\\
10 & Olympus E-M10 Mark II & $3456 \times 4608$ & Four Thirds & 17.3$\times$ 13 mm$^2$ & Bayer Filter & Interchangeable \\
11 & Sigma Sd Quattro & $3616 \times 5424$ & Foveon X3 & 23.4$\times$ 15.5 mm$^2$ & NA & Interchangeable \\
12 & Sony Alpha 68 & $4000 \times 6000$ & APS-C& 23.5$\times$15.6 mm$^2$ & Bayer Filter & Interchangeable \\
13 & Sony RX100\_1 & $3648 \times 5472$ & 1'' & 13.2 $\times$ 8.8 mm$^2$ & Bayer Filter & Fixed \\
14 & Sony RX100\_2 & $3648 \times 5472$ & 1'' & 13.2$\times$ 8.8 mm$^2$ & Bayer Filter & Fixed \\ \hline
\end{tabular}
\end{center}
\end{table*}
\subsection{The selection of cameras}
The images from the Warwick Image Forensics Dataset are captured by 14 digital cameras. The details and the technical specifications of the cameras are shown in Table \ref{tab:cameras}. The primary goal of this dataset is helping the digital forensic community to develop better understanding of the impacts from both camera exposure parameter settings and multi-frame computational photography algorithms, especially HDR imaging, on device fingerprinting. The choice of using digital cameras instead of smartphone cameras in this dataset allows us to have better control on camera exposure parameter settings during the image capturing process. And with these fine controls, the images captured are suitable for different HDR algorithms, whether they are using images of the same or different exposures to produce HDR images. The 14 cameras are from 11 different models and cover a good range of major camera manufacturers. Also, the 14 cameras show good diversity of different image sensor formats with the smallest sensor of comparable size to the sensors used on smartphones cameras.
\subsection{Image Acquisition}
The images from this dataset can be categorized into the following three classes:
\begin{itemize}
\item Flatfield images
\item SDR images
\item HDR-ready SDR images
\end{itemize}
\hspace{0.3cm}  The \textit{flatfield images} are mainly for reference SPN extraction. For each camera, 100 flatfield images are captured by taking photos of a flat blue board with the lenses adjusted to be out of focus. For each image shot, the camera is set to its lowest ISO speed to reduce the amount of read-out noise in the image. The exposure metering of each shot is adjusted to normal exposure, making the images neither too dark nor too saturated.

The \textit{SDR images} in this dataset are the standard dynamic range images taken with the cameras' single-shot mode and thus cannot be used for HDR merging algorithms. These images are taken with systematic control of the cameras' ISO speed. For each camera, images are taken with the ISO speed set to be one of the following values: ISO 100, 200, 400, 800, 1600, 3200 and 6400, with the only exceptions from the two Panasonic Lumix DC-TZ90 as their ISO speeds go only up to 3200. 30 images of different scenes in different conditions are taken for each above mentioned ISO speed on each camera. For each image shot, with the camera's ISO speed set, we enable the camera's \textit{Program Mode}, allowing the camera to adjust its aperture size and exposure time automatically to allow sufficient exposure. Almost all the images from this set are taken in a hand-held style. This set of images provide good diversity in scenes as well as camera exposure parameter settings at the same time.

The \textit{HDR-ready SDR images} are the set of standard dynamic range images, which can be used with different algorithms to produce HDR images. Images of 20 different scenes are taken for this set. Different HDR algorithms may require different sets of images. For example, \cite{debevec2008recovering} uses set of images of varying exposure times and \cite{hasinoff2016burst} expects a burst shot of under-exposure images with the same exposure time, we took continuous shots of images using three different modes. The first one is using the auto exposure bracketing (AEB) function on each camera. The AEB function allows us to take continuous shots of images with varying exposure times. The second and third modes both use fast continuous shot mode to take at least 7 continuous shots of images with the same exposure. However, one set is taken at normal exposure and the other is taken as under-exposed, usually by 1 or 2 stops measured by the cameras' exposure metering system. An example of the images taken with these three modes are shown in Fig. 1. Furthermore, to increase the diversity in exposure parameter settings, we systematically repeat these three modes with cameras set to 7 different ISO speeds as mentioned above. Thus, for each camera, more than 120 images of the same scene with various camera parameter settings are taken. The 20 different scenes included in this dataset are carefully selected, covering both indoor and outdoor, day-light and night environment, still and dynamic scenes as well as objects with different texture. The images are taken with the cameras either hand-held or sat on a tripod. With such a good diversity of camera exposure parameter settings, these images can be easily adopted by different HDR imaging algorithms and be used for other camera exposure parameter setting dependent studies as well.\\
\indent For every image from our Warwick Image Forensics Dataset, both the unaltered RAW image file and the camera generated JPEG image file are available.
\begin{figure*}
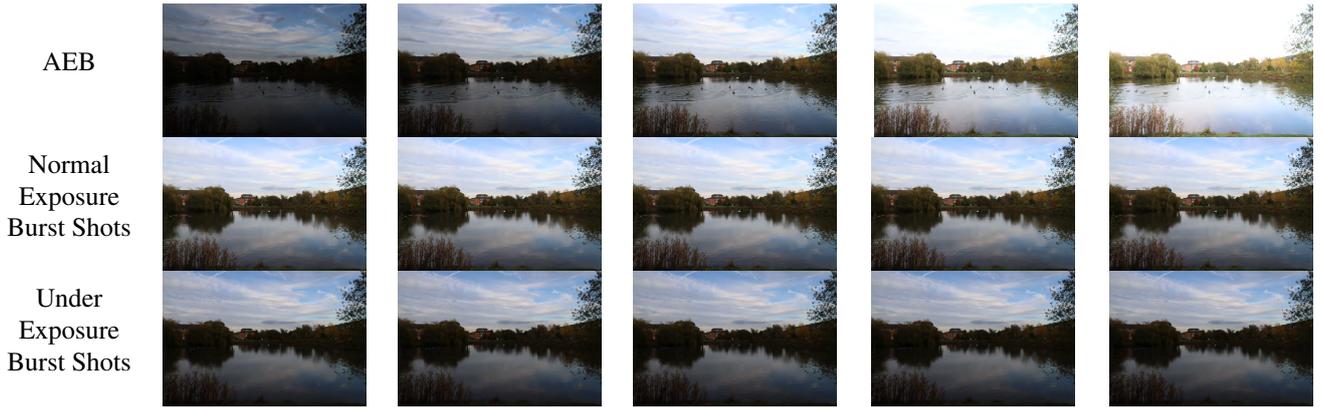

\begin{center}
\begin{tabular}{cccccc}
AEB&
\raisebox{-.5\height}{\includegraphics[width =.15\textwidth]{aeb/IMG_2095.pdf}}
 & \raisebox{-.5\height}{\includegraphics[width = .15\textwidth]{aeb/IMG_2096.pdf}} &
 \raisebox{-.5\height}{\includegraphics[width = .15\textwidth]{aeb/IMG_2094.pdf}} &
 \raisebox{-.5\height}{ \includegraphics[width = .15\textwidth]{aeb/IMG_2097.pdf}}
 & \raisebox{-.5\height}{\includegraphics[width = .15\textwidth]{aeb/IMG_2098.pdf}}
\\
\makecell{Normal\\Exposure\\Burst Shots}&
\raisebox{-.5\height}{\includegraphics[width =.15\textwidth]{burst_normal/IMG_2314.pdf}}&
\raisebox{-.5\height}{\includegraphics[width =.15\textwidth]{burst_normal/IMG_2315.pdf}}
 & \raisebox{-.5\height}{\includegraphics[width = .15\textwidth]{burst_normal/IMG_2316.pdf}} &
 \raisebox{-.5\height}{\includegraphics[width =.15\textwidth]{burst_normal/IMG_2317.pdf}}
 & \raisebox{-.5\height}{\includegraphics[width = .15\textwidth]{burst_normal/IMG_2318.pdf}}
\\
\makecell{Under\\Exposure\\Burst Shots}&
\raisebox{-.5\height}{\includegraphics[width =.15\textwidth]{burst_under/IMG_2299.pdf}}&
\raisebox{-.5\height}{\includegraphics[width =.15\textwidth]{burst_under/IMG_2300.pdf}}
 & \raisebox{-.5\height}{\includegraphics[width = .15\textwidth]{burst_under/IMG_2301.pdf}} &
 \raisebox{-.5\height}{\includegraphics[width =.15\textwidth]{burst_under/IMG_2302.pdf}}
 & \raisebox{-.5\height}{\includegraphics[width = .15\textwidth]{burst_under/IMG_2303.pdf}}
\\
\end{tabular}
\end{center}
\caption{Sample images of a scene from the \textit{HDR-ready SDR images} in Warwick Image Forensics Dataset. These images are taken by a Canon EOS 6D Mark II with ISO speed set to 100. From top to bottom, we show the images taken with three different modes. The top one uses the camera's auto exposure bracketing (AEB) function and the following two rows are shots with consistent exposure time within each row. The middle row has normal exposure and the images in the bottom row are under exposed by 1 stop measured by the cameras exposure metering system. Due to the limit of space, we only show a portion of the images taken with three modes at ISO 100.}
\vspace{-.5cm}
\end{figure*}
\section{Experimental Evaluations}
\label{sec:exp}
\begin{figure}
\begin{center}
\includegraphics[width=.37\textwidth]{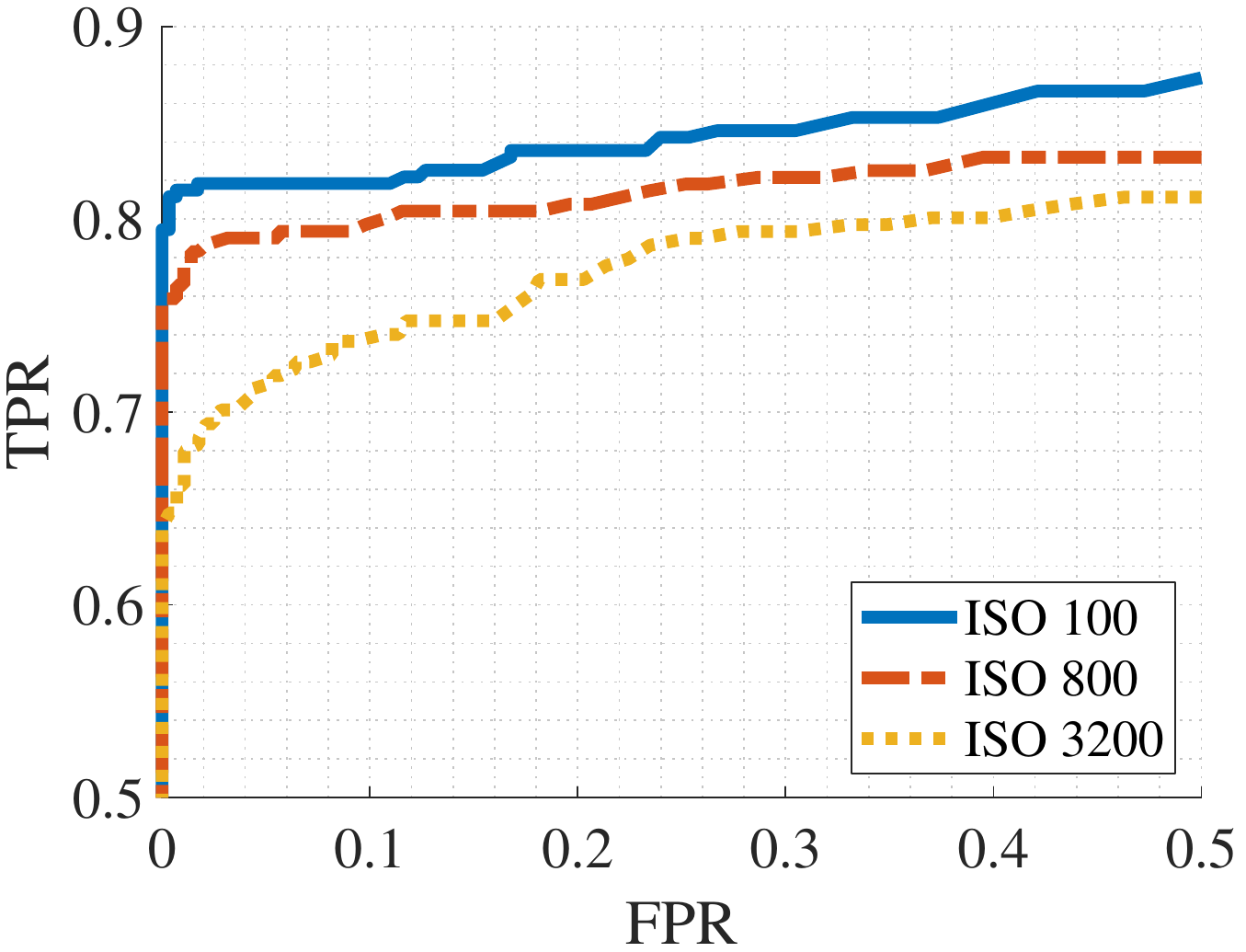}
\caption{The ROC curves of source camera identification \cite{1634362} on SDR images with ISO speed 100, 800 and 3200}
\label{fig:ROC}
\end{center}
\vspace{-.5cm}
\end{figure}
\begin{figure*}
\begin{center}

\begin{tabular}{ccc}
ISO 100 & ISO 800 & ISO 3200\\
\includegraphics[width=.24\textwidth]{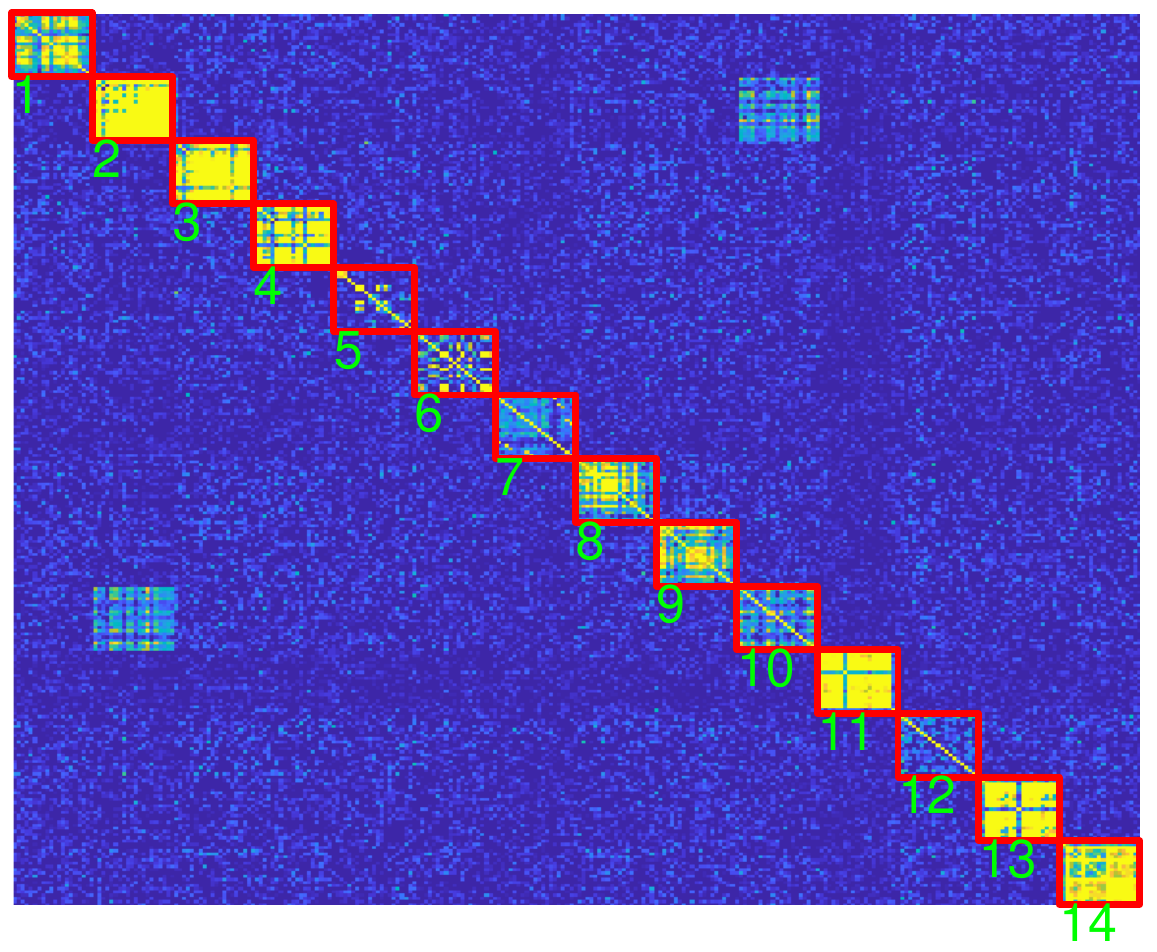} & \includegraphics[width=.24\textwidth]{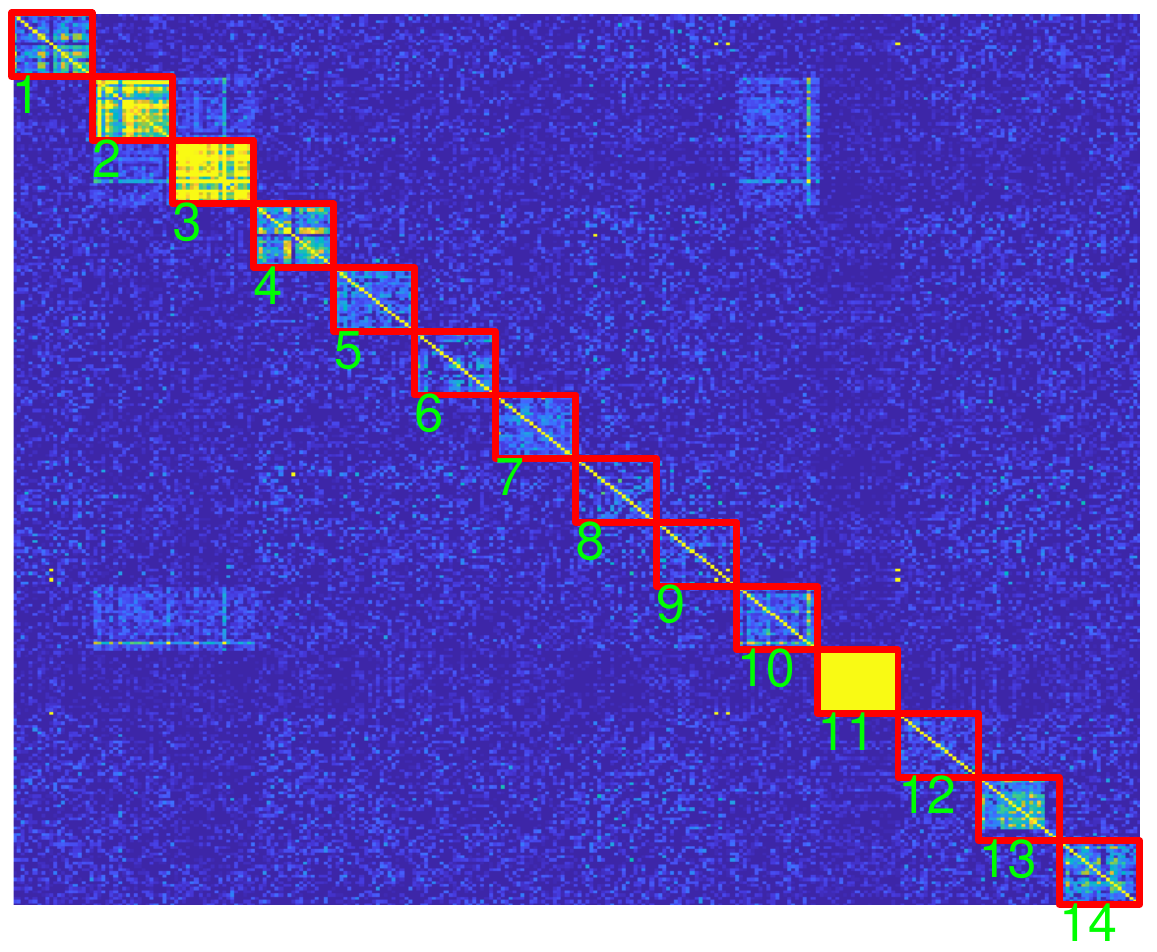}
 & \includegraphics[width=.26\textwidth]{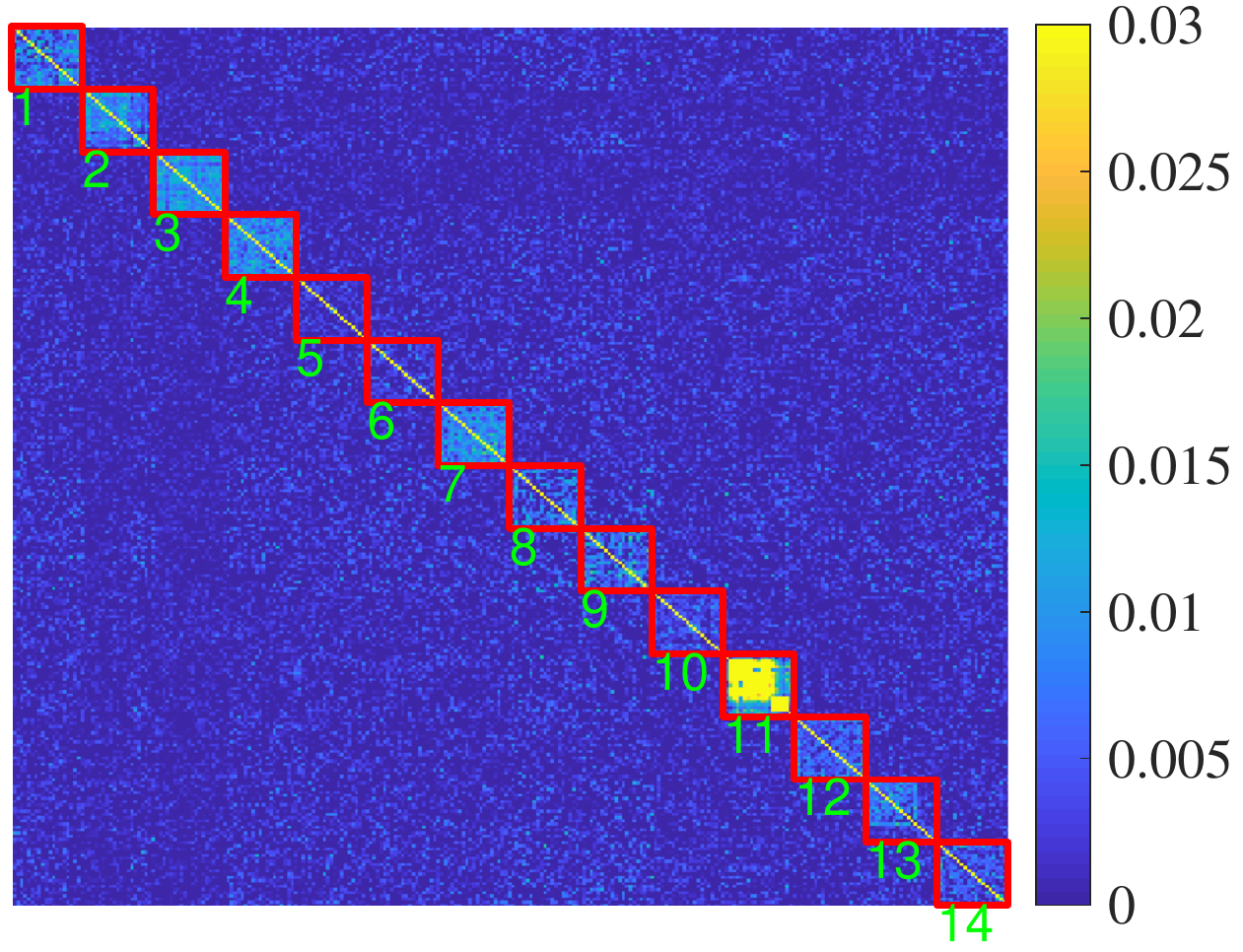}

\end{tabular}
\end{center}
\caption{Correlation matrices for the pairwise correlations between SDR images of ISO speed 100, 800 and 3200 }
\label{fig:corrMat}
\vspace{-.5cm}
\end{figure*}
In this section, we conduct experimental evaluations on SPN-based source camera identification and clustering's performance on the Warwick Image Forensics Dataset. In particular, we will show how the performance varies by using images of different ISO speeds for the tests.\\
\indent For source camera identification, from each camera, we extract the reference SPNs from 100 flatfield JPEG images using the BM3D de-noising algorithm \cite{BM3D2007}. The extracted reference SPNs are processed by a spectrum equalizer from \cite{lin2015preprocessing} to remove unwanted artefacts. We test the performance of source camera identification method from \cite{1634362} on the SDR images from the dataset. For each image, we crop a region of $512 \times 512$ pixels from its center to extract the noise residual and compute the correlations with the corresponding pixels from the reference SPNs. The receiver operator characteristics (ROC) curves for the method on images of ISO speed 100, 800 and 3200 are shown in Fig. \ref{fig:ROC}. Apparently, as the ISO speed gets higher, smaller under curve area is observed indicating worse performance.

Fig. \ref{fig:corrMat} shows the correlation matrices of pairwise correlations between noise residuals extracted from SDR images of ISO speed 100, 800 and 3200. On the plots, we use red squares to highlight the intra-class correlations belonging to each camera, marked by the number which follows the order in Table \ref{tab:cameras}. The three color-maps follow the same color scheme as shown in the bar on the right. The cluster structures in each plot become less clear as the ISO speed gets larger. And unsurprisingly, the clustering performance show the same trend with smaller F1 score for the higher ISO speed. By applying the method from \cite{Li2017}, we have F1 score of $84.33\%$, $82.86\%$ and $80.13\%$ for ISO speed 100, 800 and 3200, respectively.

All experiments mentioned above prove that different camera exposure settings have different levels of impact on the quality of SPN and the forensic analyses, which need to be considered in forensic research and real-world investigations. Therefore, it is important to include images of diverse camera parameter settings in the image datasets in order to facilitate future research.

\section{Conclusion}
\label{sec:con}
In this paper, we demonstrated the impact of camera exposure parameter settings like ISO speed on the quality of SPN and the importance of having an image dataset that can facilitate future research into the development of better solutions to deal with this impact. We presented the Warwick Image Forensics Dataset, a novel forensic image dataset consisting of more than 58,600 images, captured with special attentions to exposure parameter settings. The images are from 14 different digital cameras. The good diversity of camera parameter settings allows studies on different exposure parameters' impact on device fingerprinting to be carried out on this dataset. With the diverse ways of taking these images, they can easily be used by different multi-frame computational photography algorithms including HDR imaging. Thus, HDR image related studies in device fingerprinting can be carried out using this dataset as well. In addition, the dataset can also be used for other studies like steganalysis. Thus, we believe it is beneficial for the digital forensic community with the dataset released as an open-source.
\section*{Acknowledgment}
This work is supported by the EU Horizon 2020 Marie Sklodowska-Curie Actions through the project entitled Computer Vision Enabled Multimedia Forensics and People Identification (Project No. 690907, Acronym: IDENTITY)
\bibliography{refs.bib}

\begin{thebibliography}{10}
\providecommand{\url}[1]{#1}
\csname url@samestyle\endcsname
\providecommand{\newblock}{\relax}
\providecommand{\bibinfo}[2]{#2}
\providecommand{\BIBentrySTDinterwordspacing}{\spaceskip=0pt\relax}
\providecommand{\BIBentryALTinterwordstretchfactor}{4}
\providecommand{\BIBentryALTinterwordspacing}{\spaceskip=\fontdimen2\font plus
\BIBentryALTinterwordstretchfactor\fontdimen3\font minus
  \fontdimen4\font\relax}
\providecommand{\BIBforeignlanguage}[2]{{%
\expandafter\ifx\csname l@#1\endcsname\relax
\typeout{** WARNING: IEEEtran.bst: No hyphenation pattern has been}%
\typeout{** loaded for the language `#1'. Using the pattern for}%
\typeout{** the default language instead.}%
\else
\language=\csname l@#1\endcsname
\fi
#2}}
\providecommand{\BIBdecl}{\relax}
\BIBdecl

\bibitem{1634362}
J.~{Lukas}, J.~{Fridrich}, and M.~{Goljan}, ``Digital camera identification
  from sensor pattern noise,'' \emph{IEEE Transactions on Information Forensics
  and Security}, vol.~1, no.~2, pp. 205--214, June 2006.

\bibitem{li2010source}
C.-T. Li, ``Source camera identification using enhanced sensor pattern noise,''
  \emph{IEEE Transactions on Information Forensics and Security}, vol.~5,
  no.~2, pp. 280--287, 2010.

\bibitem{chen2008}
M.~Chen, J.~Fridrich, M.~Goljan, and J.~Luk{\'a}s, ``Determining image origin
  and integrity using sensor noise,'' \emph{IEEE Transactions on Inforamtion
  Forensics and Security}, vol.~3, no.~1, pp. 74--90, 2008.

\bibitem{5934587}
C.-T. {Li} and Y.~{Li}, ``Color-decoupled photo response non-uniformity for
  digital image forensics,'' \emph{IEEE Transactions on Circuits and Systems
  for Video Technology}, vol.~22, no.~2, pp. 260--271, Feb 2012.

\bibitem{caldelli2010fast}
R.~Caldelli, I.~Amerini, F.~Picchioni, and M.~Innocenti, ``Fast image
  clustering of unknown source images,'' in \emph{2010 IEEE International
  Workshop on Information Forensics and Security}.\hskip 1em plus 0.5em minus
  0.4em\relax IEEE, 2010, pp. 1--5.

\bibitem{Li2017}
C.-T. Li and X.~Lin, ``A fast source-oriented image clustering method for
  digital forensics,'' \emph{EURASIP Journal on Image and Video Processing:
  Special Issues on Image and Video Forensics for Social Media analysis},
  vol.~1, pp. 69--84, Oct. 2017.

\bibitem{8753503}
R.~{Rouhi}, F.~{Bertini}, D.~{Montesi}, X.~{Lin}, Y.~{Quan}, and C.-T. {Li},
  ``Hybrid clustering of shared images on social networks for digital
  forensics,'' \emph{IEEE Access}, vol.~7, pp. 87\,288--87\,302, 2019.

\bibitem{gloe2010dresden}
T.~Gloe and R.~B{\"o}hme, ``{The `Dresden Image Database' for Benchmarking
  Digital Image Forensics},'' \emph{J. of Digital Forensic Practice}, vol.~3,
  no. 2-4, pp. 150--159, 2010.

\bibitem{Shullani2017}
D.~Shullani, M.~Fontani, M.~Iuliani, O.~A. Shaya, and A.~Piva, ``Vision: a
  video and image dataset for source identification,'' \emph{EURASIP Journal on
  Information Security}, vol. 2017, no.~1, p.~15, Oct 2017.

\bibitem{reinhard2010high}
E.~Reinhard, W.~Heidrich, P.~Debevec, S.~Pattanaik, G.~Ward, and K.~Myszkowski,
  \emph{High dynamic range imaging: acquisition, display, and image-based
  lighting}.\hskip 1em plus 0.5em minus 0.4em\relax Morgan Kaufmann, 2010.

\bibitem{8451688}
L.~Lin, W.~Chen, Y.~Wang, S.~Reinder, Y.~Guan, J.~Newman, and M.~Wu, ``{The
  Impact of Exposure Settings in Digital Image Forensics},'' in \emph{2018 25th
  IEEE International Conference on Image Processing (ICIP)}, Oct 2018, pp.
  540--544.

\bibitem{mann1994beingundigital}
S.~Mann and R.~Picard, ``Being 'undigital’ with digital cameras,'' \emph{MIT
  Media Lab Perceptual}, vol.~1, p.~2, 1994.

\bibitem{debevec2008recovering}
P.~E. Debevec and J.~Malik, ``Recovering high dynamic range radiance maps from
  photographs,'' in \emph{ACM SIGGRAPH 2008 classes}.\hskip 1em plus 0.5em
  minus 0.4em\relax ACM, 2008, p.~31.

\bibitem{hasinoff2016burst}
S.~W. Hasinoff, D.~Sharlet, R.~Geiss, A.~Adams, J.~T. Barron, F.~Kainz,
  J.~Chen, and M.~Levoy, ``Burst photography for high dynamic range and
  low-light imaging on mobile cameras,'' \emph{ACM Transactions on Graphics
  (TOG)}, vol.~35, no.~6, p. 192, 2016.

\bibitem{eilertsen2017hdr}
G.~Eilertsen, J.~Kronander, G.~Denes, R.~K. Mantiuk, and J.~Unger, ``{HDR}
  image reconstruction from a single exposure using deep cnns,'' \emph{ACM
  Transactions on Graphics (TOG)}, vol.~36, no.~6, p. 178, 2017.

\bibitem{schaefer2003ucid}
G.~Schaefer and M.~Stich, ``{UCID}: An uncompressed color image database,'' in
  \emph{Storage and Retrieval Methods and Applications for Multimedia 2004},
  vol. 5307.\hskip 1em plus 0.5em minus 0.4em\relax International Society for
  Optics and Photonics, 2003, pp. 472--480.

\bibitem{dang2015raise}
D.-T. Dang-Nguyen, C.~Pasquini, V.~Conotter, and G.~Boato, ``{RAISE}: A raw
  images dataset for digital image forensics,'' in \emph{Proceedings of the 6th
  ACM Multimedia Systems Conference}.\hskip 1em plus 0.5em minus 0.4em\relax
  ACM, 2015, pp. 219--224.

\bibitem{galdi2019socrates}
C.~Galdi, F.~Hartung, and J.-L. Dugelay, ``Socrates: A database of realistic
  data for source camera recognition on smartphones,'' in \emph{Proceedings of
  International Conference Pattern Recognition Applications and Methods}, 2019,
  pp. 19--21.

\bibitem{8760241}
H.~{Tian}, Y.~{Xiao}, G.~{Cao}, Y.~{Zhang}, Z.~{Xu}, and Y.~{Zhao}, ``Daxing
  smartphone identification dataset,'' \emph{IEEE Access}, vol.~7, pp.
  101\,046--101\,053, 2019.

\bibitem{shaya2018new}
O.~Shaya, P.~Yang, Y.~Ni, R.and~Zhao, and A.~Piva, ``A new dataset for source
  identification of high dynamic range images,'' \emph{Sensors}, vol.~18,
  no.~11, p. 3801, 2018.

\bibitem{BM3D2007}
K.~Dabov, A.~Foi, V.~Katkovnik, and K.~Egiazarian, ``Image denoising by sparse
  3-d transform-domain collaborative filtering,'' \emph{IEEE Transactions on
  Image Processing}, vol.~16, no.~8, pp. 2080--2095, Aug 2007.

\bibitem{lin2015preprocessing}
X.~Lin and C.-T. Li, ``{Preprocessing Reference Sensor Pattern Noise via
  Spectrum Equalization},'' \emph{IEEE Transactions on Inforamtion Forensics
  and Security}, vol.~11, no.~1, pp. 126--140, 2016.

\end{thebibliography}

\end{document}